\documentclass{article}

\usepackage{microtype}
\usepackage{graphicx}
\usepackage{subcaption}
\usepackage[pdflang=en-US]{hyperref}
\usepackage{booktabs} 
\usepackage{hyperref}
\usepackage{amsmath}
\usepackage{amsmath,amsfonts,amsthm,bm} 
\DeclareMathOperator*{\argmax}{argmax}
\usepackage{amssymb}
\usepackage{dsfont}
\newcommand{\R}{\mathbb{R}}
\newcommand{\I}{\mathbb{I}}
\usepackage{xcolor}

\usepackage[accepted]{icml2021}

\icmltitlerunning{PHEW : Constructing Sparse Networks that Learn Fast and Generalize Well without Training Data}

\begin{document}

\twocolumn[
\icmltitle{PHEW : Constructing Sparse Networks that Learn Fast\\ and Generalize Well Without Training Data}


\begin{icmlauthorlist}
\icmlauthor{Shreyas Malakarjun Patil}{to}
\icmlauthor{Constantine Dovrolis}{to}
\end{icmlauthorlist}

\icmlaffiliation{to}{School of Computer Science, Georgia Institute of Technology, USA}

\icmlcorrespondingauthor{Constantine Dovrolis}{constantine@gatech.edu}

\icmlkeywords{Deep Learning, Architectures, Sparse Networks}
\vskip 0.3in]

\printAffiliationsAndNotice{} 

\begin{abstract}
Methods that sparsify a network at initialization  are important in practice because they greatly improve the efficiency of both learning and inference. Our work is based on a recently proposed decomposition of the Neural Tangent Kernel (NTK) that has decoupled the dynamics of the training process into a data-dependent component and an architecture-dependent kernel – the latter referred to as Path Kernel. That work has shown how to design sparse neural networks for faster convergence, without any training data, using the Synflow-L2 algorithm. We first show that even though Synflow-L2 is optimal in terms of convergence, for a given network density, it results in sub-networks with ``bottleneck'' (narrow) layers – leading to poor performance as compared to other data-agnostic methods that use the same number of parameters. Then we propose a new method to construct sparse networks, without any training data, referred to as Paths with Higher-Edge Weights (PHEW). PHEW is a probabilistic network formation method based on biased random walks that only depends on the initial weights. It has similar path kernel properties as Synflow-L2 but it  generates much wider layers, resulting in better generalization and performance. 
PHEW achieves significant improvements over the data-independent SynFlow and SynFlow-L2 methods at a wide range of network densities.
\end{abstract}

\section{Introduction}

Generating sparse neural networks through pruning has recently led to a major reduction in the number of parameters, while having minimal loss in performance. Conventionally, pruning methods operate on pre-trained networks. Generally, such methods use an edge scoring mechanism for eliminating the less important connections. Popular scoring mechanisms include weight magnitudes \cite{han2015learning,janowsky1989pruning,park2020lookahead}, loss sensitivity with respect to units \cite{mozer1989skeletonization} and with respect to weights \cite{karnin1990simple},  Hessian \cite{lecun1990optimal,hassibi1993second}, and first and second order Taylor expansions \cite{molchanov2016pruning,molchanov2019importance}. More recent approaches use  more sophisticated variants of these scores \cite{han2015deep,guo2016dynamic,carreira2018learning,yu2018nisp,dong2017learning,guo2016dynamic}.

Further analysis of pruning has shown the existence of sparse subnetworks at initialization which, when trained, are capable of matching the performance of the fully-connected network \cite{frankle2018lottery,frankle2019stabilizing,liu2018rethinking,frankle2020linear}. 
However, identifying such ``winning ticket'' networks requires expensive training and pruning cycles.
More recently, SNIP \cite{lee2018snip}, \cite{you2019drawing} and GraSP \cite{wang2020picking} showed that it is possible to find ``winning tickets'' prior to training -- but still having access to at least some training data to compute initial gradients.  
Furthermore, other work has shown that such subnetworks generalize well across datasets and tasks \cite{morcos2019one}. 

Our goal is to identify sparse subnetworks that perform almost as well as the fully connected network without {\em any} training data. 
The closest methods that tackle the same problem with our work are SynFlow \cite{tanaka2020pruning} and SynFlow-L2 \cite{anonymous2021a}.
The authors of \cite{tanaka2020pruning} introduced the concept of ``layer collapse'' in pruning -- the state when all edges in a layer are eliminated while there are edges in other layers that can be pruned. 
They also proved that iterative pruning based on positive gradient-based scores avoids layer collapse and introduced an iterative algorithm (SynFlow) and a loss function that conserves information flow and avoids layer collapse. 

A branch of recent work focuses on  the convergence and generalization properties of deep neural networks using linear approximations of the training dynamics \cite{jacot2018neural,lee2019wide,arora2019fine}. 
Under the infinite-width assumption,
\cite{jacot2018neural} showed how to predict at initialization output changes during training using the Neural Tangent Kernel (NTK). 
More recently,  \cite{anonymous2021a} decomposed the NTK into two factors: one that is only data-dependent, and another that is only architecture-dependent. 
This decomposition decoupled the effects of network architecture (including sparsity and selection of initial weights) from the effect of training data on convergence.
The architecture-dependent factor can be thought of as the ``path covariance'' of the network and is referred to as {\em Path Kernel}.
The authors of \cite{anonymous2021a}  show that the training convergence of a network can be accurately predicted using the {\em path kernel trace}.
That work concluded with a pruning algorithm ({\em SynFlow-L2}) that designs sparse networks with maximum path kernel trace -- aiming to optimize at least the architectural component of the network's convergence speed. 

In this work, we first show that even though SynFlow and Synflow-L2 are optimal in terms of convergence for a given network density, they result in sub-networks with ``bottleneck layers'' (very small width) – leading to poor performance as compared to other data-agnostic methods that use the same number of parameters. 
This issue is observed even at moderate density values. 
This is expected given the recent results of \cite{golubeva2020wider}, for instance, showing that increasing the width of sparse networks, while keeping the number of parameters constant, generally improves performance.

We then present a method,
referred to as {\em PHEW (Paths with Higher Edge Weights)}, which aims to achieve the best of both worlds: {\em high path kernel trace for fast convergence, and large network width for better generalization performance.} 
Given an unpruned initialized network, and a target number of learnable parameters, PHEW selects a set of input-output paths that are conserved in the network and it prunes every remaining connection. 
The selection of the conserved paths is based strictly on their initial weight values -- and not on any training data. 
Further, PHEW induces randomness into the path selection process using random walks biased towards higher weight-magnitudes.
The network sparsification process does not require any data, and the pruned network needs to be trained only once.

We show that selecting paths with higher edge weights forms sub-networks that have higher path kernel trace than uniform random walks -- close to the trace obtained through SynFlow-L2.
We also show that the use of random walks results in sub-networks having high per-layer width -- similar to that of unpruned networks.
Further, PHEW avoids layer-collapse by selecting and conserving input-output paths instead of individual units or connections.
We compare the performance of PHEW against several pruning before-training methods and show that PHEW achieves significant improvements over SynFlow and SynFlow-L2. 
Additionally, we conduct a wide range of ablation studies to evaluate the efficacy of PHEW.

 \begin{figure*}
\centering
        \centering
         \begin{subfigure}[b]{0.24\textwidth}
             \centering
             \includegraphics[width=\textwidth]{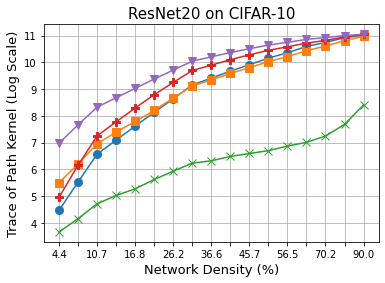}
         \end{subfigure}
           \centering
         \begin{subfigure}[b]{0.24\textwidth}
             \centering
             \includegraphics[width=\textwidth]{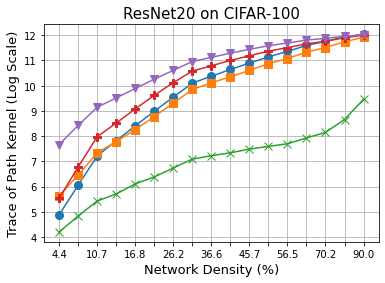}
         \end{subfigure}
         \centering
         \begin{subfigure}[b]{0.24\textwidth}
             \centering
             \includegraphics[width=\textwidth]{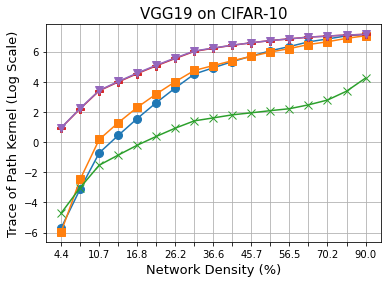}
         \end{subfigure}
           \centering
         \begin{subfigure}[b]{0.24\textwidth}
             \centering
             \includegraphics[width=\textwidth]{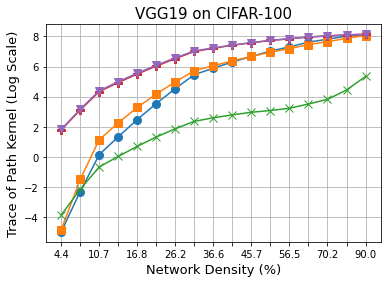}
         \end{subfigure}
           \centering
         \begin{subfigure}[b]{0.48\textwidth}
             \centering
             \includegraphics[width=\textwidth]{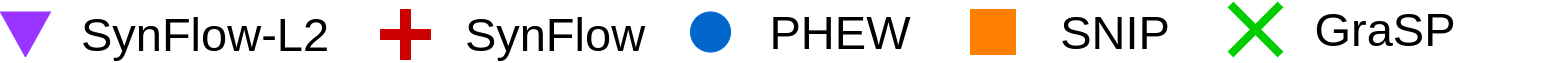}
         \end{subfigure}
    \caption{\textbf{Comparison of the path kernel trace of sparse networks obtained using various pruning methods as well as PHEW.}}
    \label{pktracefigure}
\end{figure*}

\section{Background}

\subsection{Neural Tangent Kernel}

Let $\bm{(\mathcal{X},\mathcal{Y})}$ denote the training examples,  $\bm{\mathcal{L}}$ the loss function, and $\bm{f}(\bm{\mathcal{X}},\bm{\mathcal{\theta}}) \in \R^{NK}$  the network's output, where $N$ is the number of examples and $K$ is the output dimension.
Under the gradient flow assumption, and denoting the learning rate by $\eta$, the output of the network at time $t$ can be approximated using the first-order Taylor expansion,
\begin{equation}\label{ntk}
    \bm{f}(\bm{\mathcal{X}},\bm{\mathcal{\theta}}_{t+1}) = \bm{f}(\bm{\mathcal{X}},\bm{\mathcal{\theta}}_{t}) - \eta \,\bm{\Theta}_t(\bm{\mathcal{X}},\bm{\mathcal{X}}) \, \nabla_{\bm{f}} \bm{\mathcal{L}}
\end{equation}
where the matrix $\bm{\Theta}_t(\bm{\mathcal{X}},\bm{\mathcal{X}}) = \nabla_\theta\bm{f}(\bm{\mathcal{X}},\bm{\mathcal{\theta}}_{t}) \,\nabla_\theta\bm{f}(\bm{\mathcal{X}},\bm{\mathcal{\theta}}_{t})^T$ $ \in \R^{NK\times NK}$ is the \textbf{Neural Tangent Kernel} (NTK) at time $t$
\cite{jacot2018neural}.
Under the additional assumption of infinite width, the NTK has been shown to remain constant throughout training, and it allows us to exactly predict the evolution of the network's output. 
More recent work has shown that even networks with limited width, and any depth, closely follow the NTK dynamics \cite{lee2019wide}.
For a constant NTK $\bm{\Theta}_0$, and with a mean-squared error loss, equation (\ref{ntk}) has the closed-form solution:
\begin{equation}\label{ntk1}
    \bm{f}(\bm{\mathcal{X}},\bm{\mathcal{\theta}}_{t}) = (\bm{\mathcal{I}} - e^{-\eta \bm{\Theta}_0 t})\bm{\mathcal{Y}} + e^{-\eta \bm{\Theta}_0 t} f(\bm{\mathcal{X}},\bm{\theta}_0)
\end{equation}
 Equation (\ref{ntk1}) allows us to predict the network's output given the input-output training examples, the initial weights $\bm{\theta}_0$, and the initial NTK $\bm{\Theta}_0$.
 Further, leveraging equation (\ref{ntk1}), it has been shown that the training convergence is faster in the directions that correspond to the larger NTK eigenvalues \cite{arora2019fine}. 
 This suggests that sparse sub-networks that preserve the larger NTK eigenvalues of the original network would converge faster and with higher sampling efficiency \cite{wang2020picking}.

\subsection{Path kernel decomposition of NTK}
 
More recently, an interesting decomposition of the Neural Tangent Kernel has been proposed that decouples the effects of the network architecture (and initial weights) from the data-dependent factors of the training process \cite{anonymous2021a}. We summarize this decomposition next.

Consider a neural network $\bm{f} : \R^D\rightarrow \R^K$ with ReLU activations, parametrized by $\bm{\theta} \in \R^m$.
Let $\bm{\mathcal{P}}$ be the set of all input-output paths, indexed as $p = 1,\dots,P$ (we refer to a path by its index $p$).
Let $p_i = \I\{\theta_i\in p\}$ represent the presence of edge-weight $\theta_i$ in path $p$.

 The edge-weight-product of a path is defined as the product of edge-weights present in a path, $\pi_p(\bm{\theta}) = \prod_{i=1}^m \theta_{i}^{p_i}$.
 For an input variable $\bm{x}$, the activation status of a path is, $a_p(\bm{x}) =  \prod_{\theta_i\in p} \I\{z_i>0\}$, where $z_i$ is the activation of the neuron connected to the previous layer through $\theta_i$. 
 The $k^{th}$ output of the network
 can be expressed as:
  \begin{equation}
\bm{f}^k(\bm{x},\bm{\theta}) = \sum_{i=1}^{D} \sum_{p \in \bm{\mathcal{P}}_{i\rightarrow k}} \pi_p(\bm{\theta}) \,a_p(\bm{x})\,x_i,
\end{equation} 
 where $\bm{\mathcal{P}}_{i\rightarrow k}$ is the set of paths from input unit $i$ to output unit $k$.
 We can now decompose the NTK using the chain rule:
 \begin{equation}
\begin{aligned}
 \bm{\Theta}(\bm{\mathcal{X}},\bm{\mathcal{X}})& = \nabla_{\bm{\pi}}\bm{f}(\bm{\mathcal{X}})\, \nabla_{\bm{\theta}}\bm{\pi}(\bm{\mathcal{\theta}})\, \nabla_{\bm{\theta}}\bm{\pi}(\bm{\mathcal{\theta}})^T\, \nabla_{\bm{\pi}}\bm{f}(\bm{\mathcal{X}})^T\\& = 
 \bm{J}_{\bm{\pi}}^{\bm{f}}(\bm{\mathcal{X}}) \,\bm{J}_{\bm{\theta}}^{\bm{\pi}}\, (\bm{J}_{\bm{\theta}}^{\bm{\pi}})^T \, \bm{J}_{\bm{\pi}}^{\bm{f}}(\bm{\mathcal{X}})^T\\& = \bm{J}_{\bm{\pi}}^{\bm{f}}(\bm{\mathcal{X}}) \,\bm{\Pi}_{\bm{\theta}} \,\bm{J}_{\bm{\pi}}^{\bm{f}}(\bm{\mathcal{X}})^T 
\end{aligned}
\end{equation}
The matrix $\bm{\Pi_\theta}$ is referred to as the {\bf Path Kernel} \cite{anonymous2021a}. 
The path kernel element for two paths $p$ and $p'$ is:
\begin{equation}
\bm{\Pi_\theta}(p, p') = \sum_{i=1}^m \left(\dfrac{\pi_p(\bm{\theta})}{\theta_{i}}\right)\left(\dfrac{\pi_{p'}(\bm{\theta})}{\theta_{i}}\right) p_i p'_i
\end{equation}
Note that the path kernel, $\bm{\Pi_\theta} \in \R^{P\times P}$, depends only on the network architecture and the initial weights.
On the other hand, the matrix  $\bm{J}_{\bm{\pi}}^{\bm{f}}(\bm{\mathcal{X}}) \in \R^{NK\times P}$ captures the data-dependent activations and re-weights the paths on the basis of the training data input.

\paragraph{Convergence approximation:} The eigenstructure of the NTK depends on how the eigenvectors of $\bm{J}_{\bm{\pi}}^{\bm{f}}(\bm{\mathcal{X}})$ map onto the eigenvectors of the path kernel $\bm{\Pi_\theta}$, as shown by the following result.

\textbf{Theorem 1 }\cite{anonymous2021a}: \textit{Let $\lambda_i$ be the eigenvalues of $\bm{\Theta}(\bm{\mathcal{X}},\bm{\mathcal{X}})$, $v_i$ the eigenvalues of $\bm{J}_{\bm{\pi}}^{\bm{f}}(\bm{\mathcal{X}})$ and $w_i$ the eigenvalues of $\bm{\Pi_\theta}$. Then $\lambda_i \leq v_iw_i$ and $\sum_i \lambda_i \leq \sum_i v_iw_i$}.

Given the eigenvalue decomposition of $\bm{\Theta}_0$, Theorem 1 provides an upper bound for the convergence in equation (\ref{ntk1}).
$\bm{\Theta}_0$ with eigenvalues $\lambda_i$ has the same eigenvectors as $e^{-\eta \bm{\Theta}_0 t}$ with eigenvalues $e^{-\eta \lambda_i t}$. 
Therefore, $\sum_i v_iw_i$ accurately captures the eigenvalues of $\bm{\Theta}_0$ and it can be used to predict the convergence of the training process. 
Even without any training data, the convergence can be effectively approximated from the trace of the path kernel: 
\begin{equation}\label{pkt}
    Tr(\bm{\Pi_\theta}) = \sum_i w_i=\sum_{p=1}^P \bm{\Pi_\theta}(p,p) = \sum_{p=1}^P \sum_{i=1}^m  \left(\dfrac{\pi_{p}(\bm{\theta)}}{\theta_{i}}\right)^2 p_i
\end{equation}
The authors of \cite{anonymous2021a} empirically validated the convergence predicted using the trace of the path kernel  against the actual training convergence of the network. 

The previous result has an important consequence for neural network pruning.
Given a fully connected neural network at initialization as well as the target density for a pruned network, maximizing the path kernel trace of the pruned network preserves the largest NTK eigenvalues of the original network. 
Since, the directions corresponding to the larger eigenvalues of the NTK learn faster, the sub-network obtained by maximizing the path kernel trace is also expected to converge faster and learn more efficiently.

\begin{figure*}
\centering
        \centering
         \begin{subfigure}[b]{0.24\textwidth}
             \centering
             \includegraphics[width=\textwidth]{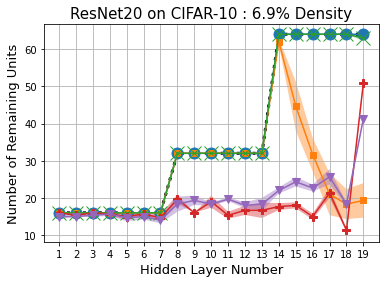}
         \end{subfigure}
           \centering
         \begin{subfigure}[b]{0.24\textwidth}
             \centering
             \includegraphics[width=\textwidth]{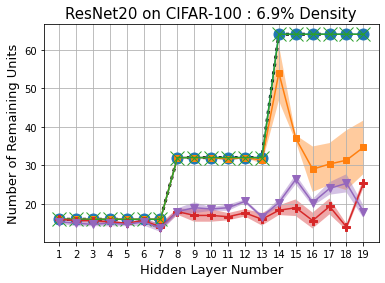}
         \end{subfigure}
         \centering
         \begin{subfigure}[b]{0.24\textwidth}
             \centering
             \includegraphics[width=\textwidth]{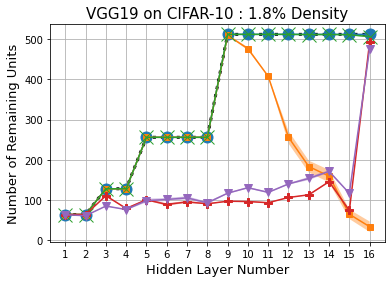}
         \end{subfigure}
           \centering
         \begin{subfigure}[b]{0.24\textwidth}
             \centering
             \includegraphics[width=\textwidth]{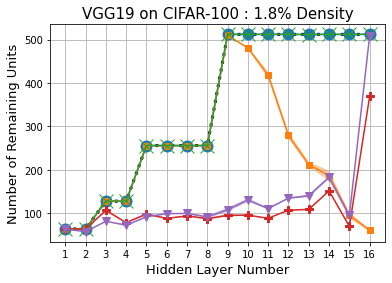}
         \end{subfigure}
           \centering
         \begin{subfigure}[b]{0.6\textwidth}
             \centering
             \includegraphics[width=\textwidth]{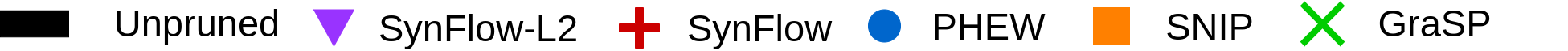}
         \end{subfigure}
    \caption{\textbf{Comparison of the number of remaining units at each layer. } The network density is selected such that the method of Magnitude-Pruning after training is able to achieve within 5\% of the unpruned network's accuracy (see text for justification).}
    \label{widthfigure}
\end{figure*}

\subsection{SynFlow-L1 and SynFlow-L2}
The path kernel framework has been applied in the design of pruning algorithms that do not require  any training data \cite{anonymous2021a,tanaka2020pruning}.
SynFlow-L2 is such an iterative pruning algorithm that removes edges (parameters) based on the following saliency function:
\begin{equation}\label{sfloss2}
S(\theta_{i}) =  \theta_{i} \odot \dfrac{\partial \mathcal{R}(\bm{\theta})}{\partial \theta_{i}^2} =\theta_{i} \odot
    \sum_{p=1}^P \left( \dfrac{\pi_p(\bm{\theta)}}{\theta_{i}} \right)^2 p_i
\end{equation}
The process of computing the previous saliency measure and eliminating edges with lower saliency is repeated until the required density is achieved.
SynFlow-L2 maximizes the trace of the path kernel,
and preserves the following data-independent loss function: 
\begin{equation}\label{sfloss}
\mathcal{R} (\bm{\theta}) = \bm{\mathds{1}}^T \left(\prod_{l=1}^{L+1} |\bm{\theta}^{[l]}|^2\right) \bm{\mathds{1}} = \sum_{p=1}^P ( \pi_p(\bm{\theta}) )^2
\end{equation}
where $|\bm{\theta}^{[l]}|^2$ is the matrix formed by the squares of the elements of the weight matrix at the $l^{th}$ layer and $L$ is the number of hidden layers.

We can also observe empirically in Figure \ref{pktracefigure} that SynFlow-L2 achieves the highest path kernel trace compared to other state-of-the-art pruning methods.

Another related pruning method is SynFlow-L1 (or simply ``SynFlow") -- proposed in \cite{tanaka2020pruning}. SynFlow is based on preserving the loss function
$R(\bm{\theta}) = \sum_{p=1}^P |\pi_p(\bm{\theta})|$, which is based on the edge-weight products along each input-output path (rather than their squares).


\section{Pruning for maximum path kernel trace}
In this section we analyze the resulting architecture of a sparse network that has been pruned to maximize the path kernel trace.
As discussed in the previous section,
SynFlow-L2 is a pruning method that has this objective.

Consider a network with a single hidden-layer, $\bm{f}:\R^D\rightarrow \R^D$, with $N$ hidden units and $D$ inputs and outputs. The incoming and outgoing weights $\bm{\theta}$ of each unit are initialized by sampling from $\mathcal{N}(0,1)$. Let the number of connections in the unpruned network be $M$, 
and let $m$ be the target number of connections in the pruned network, so that the resulting network density is $\rho = m/M$.

The optimization of the path kernel trace selects the $m$ out of $M$ parameters that maximize:
\begin{equation}
    \sum_{p=1}^P \sum_{i=1}^m \left(\dfrac{\pi_{p}(\bm{\theta})}{\theta_{i}}\right)^2 p_i
\end{equation}
In Appendix \ref{appmaxpkt}, we show that  {\em this maximization results in a fully-connected network in which only $n \leq N$ of the hidden-layer units remain in the pruned network -- all other units and their connections are removed. }
In other words, {\em the network that maximizes the path kernel trace has the narrowest possible hidden-layer width, given a target network density. }

We also show (Appendix \ref{sec-numpaths}) that this network architecture maximizes the number of input-output paths $P$ : \textit{Given a target density $\rho$, the maximum number of paths results when each hidden-layer has the same number of units, and the network is fully-connected.}

Intuitively, the previous results can be justified as follows, with the same weight distribution across all units of a layer, increasing the number of input-output paths $P$ results in higher path kernel trace.
To maximize $P$ with a given number of edges $m$, however, forces the pruning process to only maintain the edges of the smallest possible set of units at each layer. 
So, the networks produced by SynFlow and SynFlow-L2 tend to have narrower layers, compared to other pruning methods that do not optimize on the basis of path kernel trace.

To examine the previous claim empirically, and in the context of convolutional networks rather than MLPs, Figure \ref{widthfigure} compares
the number of remaining units at each layer after pruning, using the VGG19 and ResNet20 architectures. The target network density in these experiments is the lowest possible such that the method of Magnitude-Pruning (that can be performed only after training) achieves within 5\% of the unpruned network's accuracy. In higher densities there is still significant redundancy, while in lower densities there is no sufficient capacity to learn the given task. 
For a convolutional layer, the width of a layer is the number of channels at the output of that layer.
We find that both SynFlow and SynFlow-L2 result in pruned networks with very small width 
(``bottleneck layers'') compared to other state-of-the-art pruned networks of the same density.\footnote{In SNIP, the widest layers get pruned more aggressively as showed in \cite{tanaka2020pruning}. Due to this SNIP also leads to a decrease in width, but only at the widest layers.}\footnote{GraSP and PHEW are able to preserve the same width as the unpruned network for all the layers. The curves for GraSP (green) and PHEW (blue) overlap with the curve for the unpruned network in Figure \ref{widthfigure}.}
Further, with SynFlow and SynFlow-L2 all layers have approximately the same number of remaining units, i.e., approximately equal width.
Note that for the purposes of this analysis (Figure \ref{widthfigure}), we do not include skip connections for ResNet20 -- such connections complicate the definition of ``layer width" and paths, but without changing the main result of Figure \ref{widthfigure}. 

\begin{figure*}
\centering
        \centering
         \begin{subfigure}[b]{0.245\textwidth}
             \centering
             \includegraphics[width=\textwidth]{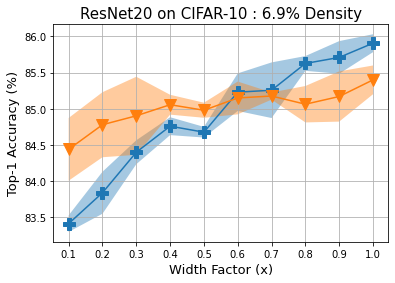}
         \end{subfigure}
         \centering
         \begin{subfigure}[b]{0.24\textwidth}
             \centering
             \includegraphics[width=\textwidth]{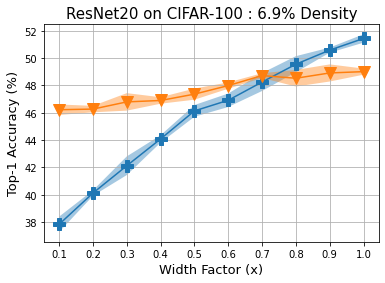}
         \end{subfigure}
          \centering
         \begin{subfigure}[b]{0.25\textwidth}
             \centering
             \includegraphics[width=\textwidth]{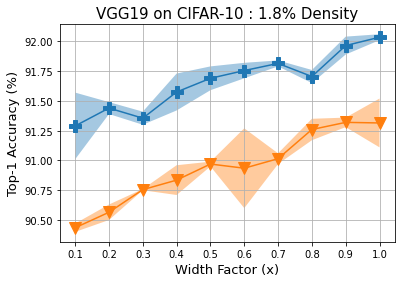}
         \end{subfigure}
         \centering
         \begin{subfigure}[b]{0.24\textwidth}
             \centering
             \includegraphics[width=\textwidth]{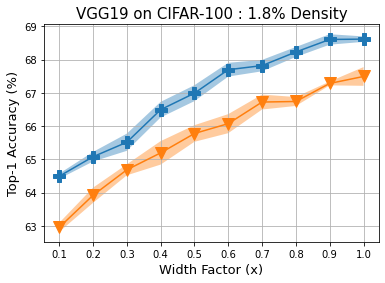}
         \end{subfigure}
           \centering
         \begin{subfigure}[b]{0.55\textwidth}
             \centering
             \includegraphics[width=\textwidth]{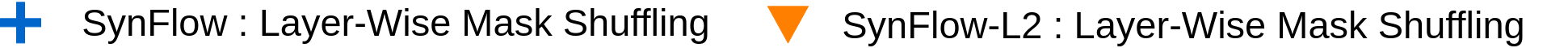}
         \end{subfigure}
    \caption{\textbf{The effect of increasing the layer width of SynFlow and SynFlow-L2 networks, while preserving the same set of parameters at each layer.} The definition of the x-axis ``Width Factor'' appears in the main text.}
    \label{111}
\end{figure*}

\subsection{Effect of network width on performance} 
Several empirical studies have been conducted to understand the effect of network width and over-parametrization on learning performance \cite{neyshabur2018role,du2018gradient,park2019effect,lu2017expressive}.
However, the previous studies do not decouple the effect of increasing width from the effect of over-parametrization.
Recently, \cite{golubeva2020wider} examined the effect of network width under a constant number of parameters.
That work conducted experiments with layer-wise random pruning.
Starting with a fully-connected network, the width of each layer is increased while keeping the number of parameters the same.
The experiments of \cite{golubeva2020wider} show that as the network width increases the performance also increases.
Further, the distance between the Gaussian kernel formed by the sparse network and the infinitely wide kernel at initialization is indicative of the network's performance.
As expected, increasing the width after a certain limit without also increasing the number of parameters will inevitably cause a  drop in both test and train accuracy because of very low per-unit connectivity (especially with random pruning).

We present similar experiments for SynFlow and SynFlow-L2 in Figure \ref{111}.
For a given network density, we first obtain the layer-wise density and number of active units that result from the previous two pruning algorithms.
We then gradually increase the number of active units by randomly shuffling the masks of each layer (so that the number of weights at each layer is preserved). 
The increase in layer width can be expressed as the fraction $x=(w'-w)/(W-w)$, where $W$ is the layer width of the unpruned network, $w$ is the layer width in the Synflow (or Synflow-L2) pruned network, and $w'\geq w$ is the layer width that results through the shuffling method described above. 
The maximum value $x=1$ results when $w'=W$.

Figure \ref{111} shows the results of these models on CIFAR-10/100 tasks using ResNet20 and VGG19. 
We can see that as the width increases so does the performance of the sparse network, even though the layer-wise number of edges is the same.
Similar results appear in the ablation studies of \cite{frankle2020pruning} using SynFlow. 
That study redistributes the edges of a layer, creating a uniform distribution across all units in the layer -- doing so increases the performance of the network (see Appendix  \ref{SF2ablations}).

\textbf{Summary:} Let us summarize the observations of this section regarding the maximization of the path kernel trace -- and the resulting decrease in network width.
Even without any training data, pruned networks that result by maximizing the path kernel trace are expected to converge faster and learn more efficiently. 
As we showed however, for a given density, such methods tend to maximize the number of input-output paths, resulting in pruned networks with very narrow layers. 
Narrow networks, however, attain lower performance as compared to wider networks of the same layer-wise density.
In the next section, we present a method that aims to achieve the best of both worlds: high path kernel trace for fast convergence, and large layer-wise  width for better generalization and learning. 



\section{The PHEW network sparsification method}

Given a weight-initialized architecture, and a target number of learnable parameters, we select a set of input-output paths that are conserved in the network -- and prune every  connection that does not appear in those paths. 
The selection of conserved paths is based strictly on their initial weights -- not on any training data. 
The proposed method is called \textit{"Paths with Higher Edge Weights"} (PHEW) because it has a bias in favor of higher weight connections.
Further, the path selection is probabilistic, through biased random walks from input units to output units. 
Specifically, the next-hop of each path, from unit $i$ to $j$ at the next layer, is taken with a probability that is proportional to the weight magnitude of the connection from $i$ to $j$.
We show that conserving paths with higher edge weight product results in higher path kernel trace.
The probabilistic nature of PHEW avoids the creation of "bottleneck layers" and leads to larger network width than methods with similar path kernel trace. 
Additionally, the procedure of selecting and conserving input-output paths completely avoids layer collapse.

In more detail,
let us initially consider a fully-connected MLP network with $L$ hidden layers and $N_l$ units at each layer (we consider convolutional networks later in this section). Suppose that the weights are initialized according to Kaiming's method \cite{he2015delving}, i.e., they are sampled from a Normal distribution in which the variance is inversely proportional to the width of each layer: $\theta_{i,j}^{l} \thicksim \mathcal{N}(0,\sigma_l^2)$, where $\sigma_l^2 = 2/N_l$. 

First, let us consider two input-output paths $u$ and $b$:  $u$
has been selected via a uniform random-walk in which the probability $Q(j,i)$ that the walk  moves from unit $i$ to unit $j$ at the next layer is the same for all $j$;
 $b$ has been selected via the following weight-biased random-walk process:
\begin{equation}\label{brw}
    Q(j,i) = \dfrac{|\theta(j,i)|}{\sum_{j} |\theta(j,i)|}
\end{equation}
where $\theta(j,i)$ is the weight of the connection from $i$ to $j$.

In Appendix \ref{phewpktrace} we show that the biased-walk path $b$ contributes more in the path kernel trace than path $u$:
\begin{equation}
    \mathbb{E}[\bm{\Pi_\theta}(b,b)] = 2^{L} \times \mathbb{E}[\bm{\Pi_\theta}(u,u)]
\end{equation}
 As the number of hidden layers $L$ increases the ratio between the two terms becomes exponentially higher. 
 On the other hand, as the layer's width increases the ratio of two values remains the same. 
 The reason that PHEW paths result in higher path kernel trace, compared to the same number of uniformly chosen paths, is that the former tend to have higher edge weights, and thus higher $\pi_p(\bf{\theta})$ values (see Equation~\ref{pkt}).
 Empirically, Figure \ref{pktracefigure} shows that PHEW achieves a path kernel trace greater than or equal to SNIP and GraSP, and close to the upper bound of SynFlow-L2.

If the PHEW paths were chosen deterministically (say in a greedy manner, always taking the next hop with the highest weight) the path kernel trace would be slightly higher but the resulting network would have "bottlenecks" at the few units that have the highest incoming weights. 
PHEW avoids this problem by introducing randomness in the path selection process. 
Specifically, in Appendix~\ref{app:width} we show that the expected number of random walks through each unit of a layer $l$ is $W/N_l$, where $W$ is the required number of walks to achieve the target network density. Thus, as long as $W>N_l$, every unit is expected to be traversed by at least one walk -- and thus every unit of that layer is expected to be present in the sparsified network. 

This is very different than the behavior of SynFlow or SynFlow-L2, in which the width of several layers in the pruned network is significantly reduced. 
Empirically, Figure \ref{widthfigure} confirms that PHEW achieves the larger per-layer width, compared to SynFlow and SynFlow-L2. Additionally, the per-layer width remains the same as the width of the original unpruned network.

\textbf{Layer-Collapse: }
Layer collapse is defined as a network state in which all edges of a specific layer are eliminated, while there are still connections in other layers \cite{tanaka2020pruning}.
Layer collapse causes a disruption of information flow through the sparse network making the network untrainable.
SynFlow and SynFlow-L2 have been shown to avoid layer collapse by iteratively computing gradient based importance scores and pruning \cite{tanaka2020pruning}. PHEW also avoids layer collapse due to its path-based selection and conservation process. Even a single input-output path has one connection selected at each layer, and so it is impossible for PHEW networks to undergo layer collapse.

\begin{figure*}
\centering
        \centering
         \begin{subfigure}[b]{0.32\textwidth}
             \centering
             \includegraphics[width=\textwidth]{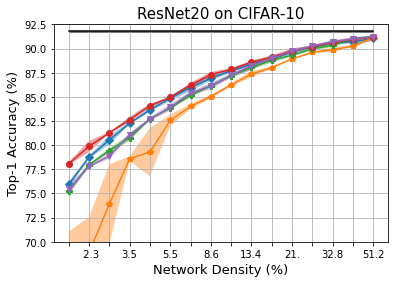}
         \end{subfigure}
         \centering
         \begin{subfigure}[b]{0.32\textwidth}
             \centering
             \includegraphics[width=\textwidth]{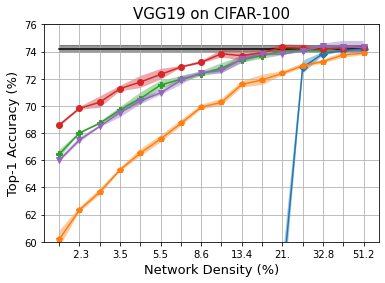}
         \end{subfigure}
         
         \centering
         \begin{subfigure}[b]{0.32\textwidth}
             \centering
             \includegraphics[width=\textwidth]{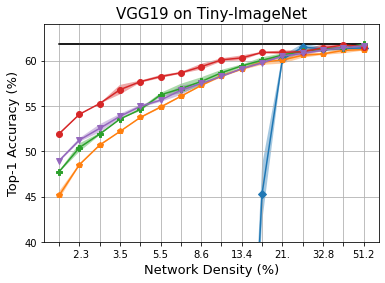}
         \end{subfigure}
         \centering
         \begin{subfigure}[b]{0.32\textwidth}
             \centering
             \includegraphics[width=\textwidth]{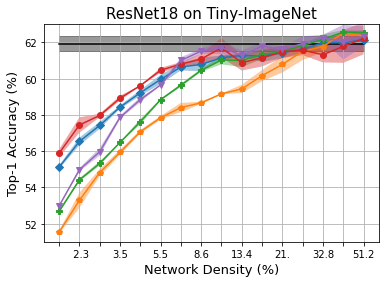}
         \end{subfigure}
          \centering
         \begin{subfigure}[b]{0.7\textwidth}
             \centering
             \includegraphics[width=\textwidth]{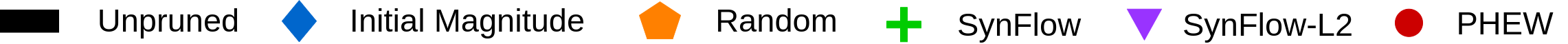}
         \end{subfigure}
    \caption{\textbf{Comparison of the Top-1 accuracy for sparse networks obtained using PHEW and other state-of-the-art data-independent baselines.} The mean is shown as a solid line while the standard deviation is shown as shaded area.}
    \label{results}
\end{figure*}

\subsection{Additional PHEW details}


\textbf{Balanced, bidirectional walks: }
Without any information about the task or the data, the only reasonable prior is to assume that every input unit is equally significant -- and the same for every output unit. 
For this reason, PHEW attempts to start the same number of walks from each input. And to terminate the same number of walks at each output. 

To do so,  we create paths in both directions with the same probability: forward paths from input units, and reverse paths from output units. 
The selection of the starting unit in each case is such that the number of walks that start (or terminate) at each input (or output) unit is approximately the same. 
The creation of random-walks continues until we have reached the given, target number of parameters.

\textbf{PHEW in convolutional neural networks:}
A convolutional layer takes as input a 3D-vector with $n_i$ channels and transforms it into another 3D-vector of $n_{i+1}$ channels. Each of the  $n_{i+1}$ units in a layer produces a single 2D-channel corresponding to the $n_{i+1}$ channels. A 2D channel is produced applying convolution on the input vector with $n_i$ channels, using a 3D-filter of depth $n_i$. Therefore each input from a unit at the previous layer has a corresponding 2D-kernel as one of the channels in the filter. So, even though MLPs have an individual weight per edge, convolutional networks have a 2D-kernel per edge. 

 A random-walk can traverse an edge of a convolutional network in two ways: either traversing a single weight in the corresponding 2D kernel -- or traversing the entire kernel with all its weights. 
 Traversing a single weight from a kernel conserves that edge and produces a non-zero output channel. 
 This creates sparse kernels and allows for the processing of multiple input channels at the same unit and with fewer parameters. 
 On the other hand, traversing the entire 2D-kernel that corresponds to an edge means that several other kernels will be eliminated. 
Earlier work in pruning has shown empirically the higher performance of creating sparse kernels instead of pruning entire kernels \cite{blalock2020state,liu2018rethinking}.
Therefore, {\em in PHEW we choose to conserve individual parameters during a random-walk rather than conserving entire kernels.} 

In summary, PHEW follows a two-step process in convolutional networks: first an edge (i.e., 2D-kernel) is selected using equation (\ref{brw}). Then a single weight is chosen from that kernel, randomly, with a probability that is proportional to the weight of the sampled parameter. 
We have also experimented with the approach of conserving the entire kernel, and we also present results for that case in the next section.

\section{Experimental results}

In this section we present several experiments conducted to compare the performance of PHEW against state-of-the-art pruning methods and other baselines. 
We present results both for standard image classification tasks using convolutional networks as well as an image-transformation task using MLPs (that task is described in more detail in Appendix \ref{B}).
We also conduct a wide variety of ablation studies to test the efficacy of PHEW.

\subsection{Classification comparisons}
We compare PHEW against two baselines, {\em Random pruning} and {\em Initial (Weight) Magnitude} pruning, along with four state-of-the-art algorithms: {\em SNIP} \cite{lee2018snip}, {\em GraSP} \cite{wang2020picking}, {\em SynFlow} \cite{tanaka2020pruning} and {\em SynFlow-L2} \cite{anonymous2021a}. 
We also present results for the original {\em Unpruned} network as well as for {\em Gradual Magnitude Pruning} \cite{zhu2017prune} to upper bound the network's performance for a given density. 
Because PHEW is data-independent, the most relevant comparison is with SynFlow and SynFlow-L2, which also do not require any training data. 

We present results for three networks: {\em ResNet20, VGG19} and {\em ResNet18} -- and three datasets: {\em CIFAR-10}, {\em CIFAR-100} and {\em Tiny ImageNet}. 
The network density range is chosen such that the magnitude pruning after training method (our performance upper bound) maintains a comparable accuracy to the unpruned network.
We claim that it is not important in practice to consider densities in which any pruned network would perform poorly. 
The hyper-parameters used were tuned only for the unpruned network (see Appendix \ref{id}).
We run each experiment three times for different seed values and present the mean and standard deviation of the test accuracy.

\begin{figure*}
\centering
        \centering
         \begin{subfigure}[b]{0.32\textwidth}
             \centering
             \includegraphics[width=\textwidth]{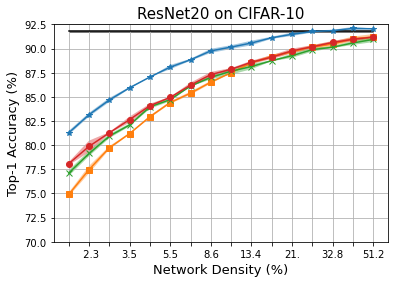}
         \end{subfigure}
         \centering
         \begin{subfigure}[b]{0.32\textwidth}
             \centering
             \includegraphics[width=\textwidth]{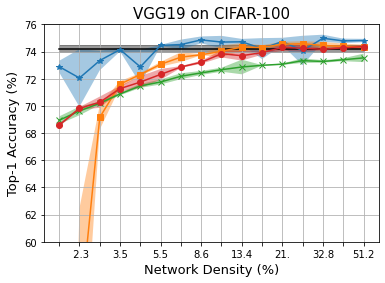}
         \end{subfigure}

         \centering
         \begin{subfigure}[b]{0.32\textwidth}
             \centering
             \includegraphics[width=\textwidth]{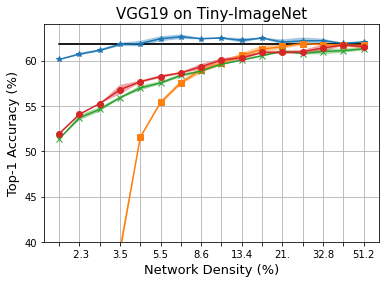}
         \end{subfigure}
         \centering
         \begin{subfigure}[b]{0.32\textwidth}
             \centering
             \includegraphics[width=\textwidth]{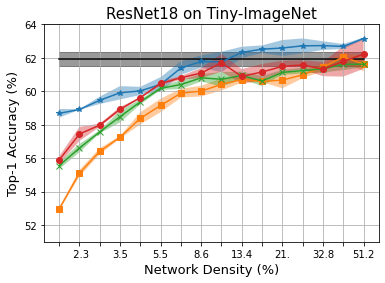}
         \end{subfigure}
          \centering
         \begin{subfigure}[b]{0.6\textwidth}
             \centering
             \includegraphics[width=\textwidth]{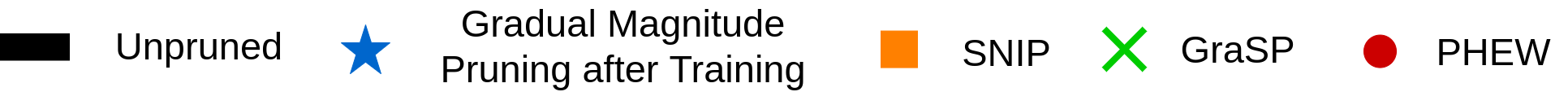}
         \end{subfigure}
    \caption{\textbf{Comparison of the Top-1 accuracy for sparse networks obtained using PHEW and other state-of-the-art data-dependent pruning methods and baselines.} The mean is shown as a solid line while the standard deviation is shown as shaded area.}
    \label{results2}
\end{figure*}
\textbf{Data-independent methods:} 
We first compare PHEW against data-independent pruning baselines random pruning, initial magnitude pruning, and methods SynFlow and SynFlowL2 in Figure~\ref{results}. 
At the density levels considered, PHEW performs better than both SynFlow and SynFlow-L2. 
We attribute this superior performance to the large per-layer width of the sparse networks obtained through PHEW.
As we showed earlier, differences in layer-wise width of sparse networks accounts for most of the performance differences, under the same number of parameters. 
Further, the performance gap increases for datasets with more classes, such as Tiny-ImageNet as compared to CIFAR-10 and CIFAR-100.
We believe that this is due to the increased complexity of the dataset and the need for larger width to learn finer and disconnected decision regions \cite{nguyen2018neural}. 



\textbf{Data-dependent methods:} Figure~\ref{results2} compares the test accuracy of PHEW networks against various data-dependent pruning algorithms. It is interesting that PHEW  outperforms in many cases  the data-dependent pruning algorithms GraSP and SNIP even though it is agnostic to the data or task. 

At higher network densities, PHEW is competitive with SNIP. Note that SNIP also utilizes paths with higher weight magnitudes \cite{anonymous2021a}.  At lower network densities,  SNIP's accuracy falls quickly while the accuracy of PHEW falls more gradually creating a significant gap. This is because SNIP eliminates units at a higher rate at the largest layer (Figure \ref{widthfigure}), which eventually leads to layer collapse \cite{tanaka2020pruning}. 

GraSP on the other hand is competitive with PHEW at lower density levels and falls short of PHEW at moderate and higher network densities. This is because GraSP maximizes gradient flow after pruning, and the obstruction to gradient flow is prominent only at lower density levels. Further, it has been shown that GraSP  performs better than SNIP only at lower density levels \cite{wang2020picking}, which is consistent with our observations. 
 \begin{figure}
\centering
        \centering
         \begin{subfigure}[b]{0.2375\textwidth}
             \centering
             \includegraphics[width=\textwidth]{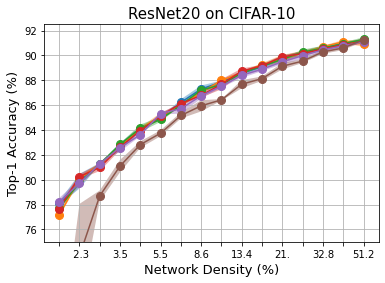}
         \end{subfigure}
         \centering
         \begin{subfigure}[b]{0.2375\textwidth}
             \centering
             \includegraphics[width=\textwidth]{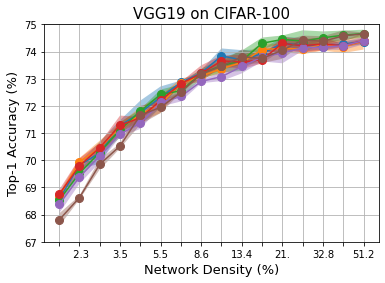}
         \end{subfigure}
          \centering
         \begin{subfigure}[b]{0.48\textwidth}
             \centering
             \includegraphics[width=\textwidth]{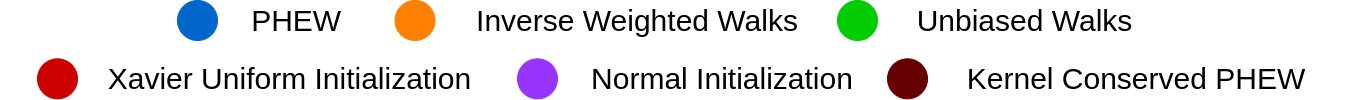}
         \end{subfigure}
    \caption{Ablation Studies: Comparison of several sparse networks obtained using variants of PHEW.}
    \label{Ablations1}
\end{figure}

PHEW is able to perform relatively well across all density levels. At lower density levels, the obstruction to the flow of gradients is avoided by conserving input-output paths. Further, by selecting higher weight magnitudes, PHEW avoids the problem of vanishing gradients.

\subsection{Ablation Studies}

In this section we conduct a wide variety of ablation studies to test PHEW's efficacy as well as to understand fully the cause behind its generalization performance. 

  \textbf{Kernel-conserved PHEW variant: }We study a PHEW variant for convolutional neural networks where instead of conserving a single weight of a kernel each time a random walk traverses that kernel, we conserve the entire kernel. This approach reduces the FLOP count immensely by eliminating the operations performed on several 2D feature maps in specific units. 
 We present the comparison for CIFAR10 and CIFAR100 in Figure \ref{Ablations1}.  Note that at moderate network densities, this kernel-conserved variant performs as well as the other methods we consider -- therefore this variant can be utilized when decreasing the FLOP count is a priority.

\textbf{Different weight initializations: }PHEW depends on initial weights, and so it is important to examine the robustness of PHEW's performance across the major weight initialization methods used in practice: Kaiming  (\cite{he2015delving}), Normal $ \mathcal{N}(0,0.1)$, and Xavier uniform (\cite{glorot2010understanding}).
Figure \ref{Ablations1} shows results with such initializations for VGG19 and ResNet20 on CIFAR10 and CIFAR100. 
 Note that PHEW's performance is quite robust across all these weight initializations.
 It is interesting that PHEW's performance is not altered by initializing all layers with the same distribution. 
 This may be due to the random walk procedure, where the probability distribution at each hop is only dependent on the initial weights of that  layer. 
 
 \textbf{Unbiased and “inverse-weight biased” random walks:} We also present two variants of PHEW: a) unbiased (uniform) walks, and b) random walks that are biased in favor of lower weight-magnitudes. The former selects the next-hop of the walk randomly, ignoring the weights, while the latter gives higher probability to lower weight-connections.

 The two variants perform similar with PHEW in terms of accuracy, as shown in Figure \ref{Ablations1}, because they create networks that have the same per-layer width with PHEW. 
 
 Their difference with PHEW is in terms of convergence speed. The reason is that they are not biased in favor of higher-weight connections, and so the resulting path kernel trace is lower than that in PHEW. Indeed, we have also confirmed empirically that the training error drops faster initially (say between epochs 1 to 20) in PHEW than in these two variants.

 
  \begin{figure}
\centering
        \centering
         \begin{subfigure}[b]{0.2375\textwidth}
             \centering
             \includegraphics[width=\textwidth]{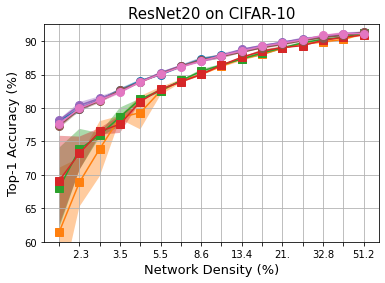}
         \end{subfigure}
           \centering
         \begin{subfigure}[b]{0.2375\textwidth}
             \centering
             \includegraphics[width=\textwidth]{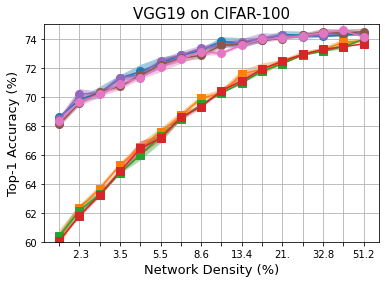}
         \end{subfigure}
           \centering
         \begin{subfigure}[b]{0.48\textwidth}
             \centering
             \includegraphics[width=\textwidth]{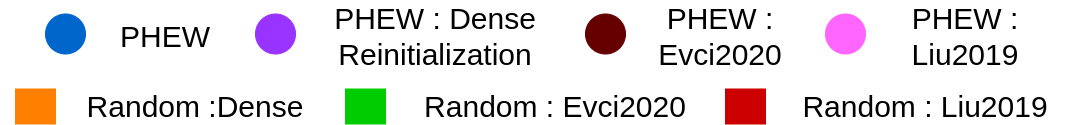}
         \end{subfigure}
    \caption{Comparison of PHEW against random pruning with sparse initialization and reinitialized PHEW networks. PHEW variants are concentrated at the upper part of plots, while random pruning variants are concentrated at the lower part.}
    \label{Ablations2}
\end{figure}

\textbf{Re-initializing PHEW Networks: } We have also conducted experiments with reinitialized sparse networks selected through PHEW. Here, the architecture of the sparse network  remains the same but the individual weights are re-sampled. This  causes the path-kernel trace to be lower while the architecture produced by PHEW is maintained. 

We present results for three reinitialization schemes, a) dense reinitialization, b) layer-wise sparse reinitialization \cite{liu2018rethinking} and c) neuron-wise sparse reinitialization \cite{evci2020gradient}.

We observe that reinitialized PHEW networks also achieve the same performance as the original PHEW network. This further confirms that the sparse architecture produced through the random walk process is sufficient to obtain PHEW's generalization performance.

In Figure \ref{Ablations2} we also compare with reinitialized randomly pruned networks, as those networks have been shown to improve performance. 
Although we see some improvements with sparse reinitialization in randomly pruned networks, the performance still falls short of PHEW  and its variants.


\section{Conclusion}

We proposed a probabilistic approach called PHEW to construct sparse networks without any training data.  
Sparse networks that result by maximizing the path kernel trace are expected to converge faster. 
We showed that, for a given density, methods that maximize the path kernel trace result in very narrow layers and lower performance compared to wider networks of the same layer-wise density.
On the other hand, conserving paths with higher edge-weight magnitudes leads to sparse networks with higher path kernel trace.
Further, introducing randomness in the path selection process preserves the layer-wise width of the unpruned network. 
Empirically, we showed that PHEW achieves significant improvements over current data-independent state-of-the-art pruning at initialization methods.



Some open questions for future research are:
1) A comparison between PHEW networks and ``winning tickets'' \cite{frankle2018lottery}, given the same number of parameters, both in terms of their convergence speed and structural properties.
2) Development of path-based network sparsification methods that can utilize a limited amount of training data to get even higher performance than PHEW. 3) How to identify path-based sparse networks that can perform as well as PHEW networks but without any training? 4) How to dynamically determine the optimal number of parameters in a sparse network at the early stages of  training -- instead of starting with a pre-determined target number of parameters? 

\section*{Acknowledgements}
This work is supported by the National Science Foundation (Award: 2039741) and by the Lifelong Learning Machines (L2M) program of DARPA/MTO (Cooperative Agreement HR0011-18-2-0019). The authors acknowledge the constructive comments given by the ICML and ICLR 2021 reviewers and by Cameron Taylor.

\bibliography{example_paper}
\bibliographystyle{icml2021}

\newpage
\appendix

\section{Proofs}\label{Proofs}

\subsection{Maximum path kernel trace} \label{appmaxpkt}

We consider a single hidden-layer network, $\bm{f}:\R^D\rightarrow \R^D$, with $N$ hidden units and $D$ inputs and outputs. The incoming and outgoing weights of each hidden unit are initialized by sampling from $\mathcal{N}(0,1)$. The number of connections in the unpruned network is $M$, 
while the target number of connections in the pruned network is $m<M$. The corresponding network density is $\rho = m/M$.

Using the notation of Section~2, the path kernel trace maximization problem is to select the $m$ edges that form a set of paths $P$ such that the following function is maximum: 
\begin{equation}
    \sum_{p=1}^P \sum_{i=1}^m \left(\dfrac{\pi_{p}(\bm{\theta})}{\theta_{i}}\right)^2 p_i, \mbox{~given~} m = \rho \times M
\end{equation}

The following lemma identifies the architecture that maximizes the previous expression, under the following two assumptions:

\textbf{1.} For simplicity, we assume that $m$ is a multiple of $2D$,  so that we can form a fully-connected network with an integer number of hidden units. 

\textbf{2.} Because all hidden units follow the same weight distribution, we assume that the sum of squares of the top-$d$ (out of $D$) outgoing weights is approximately the same for all hidden units -- and denote that constant by $K_d$. This assumption is reasonable if $D$ is large and $d \gg 1$.
Also, because the incoming weights in a hidden unit  follow the same distribution with the outgoing weights, $K_d$ also approximates the sum of squares of the top-$d$ (out of $D$) incoming weights at each hidden unit. 

\textbf{Lemma 1:} {\em The maximization of the path kernel trace results in a fully-connected network in which only $n=\frac{m}{2D}$ of the hidden-layer units remain in the pruned network -- all other units and their connections are removed. So, given a target network density, the network that maximizes the path kernel trace has the narrowest possible hidden-layer width. }

\textbf{Proof:} Consider a path $p$ defined by the set of edge-weights $\{\theta_p^{[1]}, \theta_p^{[2]} \}$ at the first and second hop, respectively.
We can re-write the optimization problem as,
\begin{equation}\label{pktracereformulation}
    \max \sum_{p=1}^P \left[(\theta_p^{[1]})^2 + (\theta_p^{[2]})^2\right], \mbox{~given~} m = \rho \times M
\end{equation}


Let us denote as $d_{in}$ the number of incoming connections of a hidden unit, and $d_{out}$ the number of outgoing connections. The total degree of that unit is 
$d_{in}+d_{out}$.
We can assume that this quantity is even for the following reason. 
The total number of edges $m$ is even, and so the total number of units with odd degree has to be even.
So, if there is a unit $j$ with odd degree, we can move a connection from another odd-degree unit $i$ to $j$ so that both $i$ and $j$ have even degree. We can repeat this step for every pair of units with odd degree until all hidden units have even degree.

Let us now consider a hidden unit $j$ with in-degree $d_{in}$ and out-degree $d_{out}$.
The path kernel trace contribution of the incoming and outgoing connections of unit $j$ is:
\begin{equation}
    \sum_{i=1}^{d_{in}} \sum_{k=1}^{d_{out}} \left[ (\theta^{[0]}(j,i))^2 + (\theta^{[1]}(k,j))^2 \right]
\end{equation}
which is equivalent with:
\begin{equation}
    d_{in} \times \sum_{k=1}^{d_{out}} (\theta^{[1]}(k,j))^2 + d_{out} \times \sum_{i=1}^{d_{in}} (\theta^{[0]}(j,i))^2
\end{equation}
To maximize this expression, we can select the top-$d_{in}$ incoming connections and the top-$d_{out}$ outgoing connections in terms of squared weights. 
Then, based on Assumption-2, the previous  expression becomes:
\begin{equation}
    d_{in} \times K_{d_{out}} + d_{out} \times K_{d_{in}}
\end{equation}

Suppose that $d_{out}\geq d_{in}+2$ (recall that $d_{out}+d_{in}$ is even and so it cannot be that $d_{out} = d_{in}+1$). Then, we can remove the connection with the lowest squared weight of the $d_{out}$ outgoing connections and include an additional incoming connection -- the one with the highest ($d_{in}+1$)-ranked squared weight). This operation will result in the following path kernel trace difference:
\begin{equation}
\begin{aligned}
    (d_{in}+1)\times K_{d_{out}-1} + (d_{out}-1)\times K_{d_{in}+1} \\ - d_{in} \times K_{d_{out}} - d_{out} \times K_{d_{in}}
\end{aligned}
\end{equation}
\begin{equation}
\begin{aligned}
    = d_{in}\times ( K_{d_{out}-1} - K_{d_{out}} ) \\ +  d_{out}\times ( K_{d_{in}+1} - K_{d_{in}}  ) \\ +  K_{d_{out}-1} - K_{d_{in}+1}
\end{aligned}
\end{equation}
This difference is always positive because $d_{out} \geq d_{in}+2$, and so $|K_{d_{out}-1} - K_{d_{out}}| \leq |K_{d_{in}+1} - K_{d_{in}}|$ and $K_{d_{out}-1} \geq K_{d_{in}+1}$. 

If $d_{in}\geq d_{out}+2$ we repeat the same process but removing an incoming connection and adding an outgoing connection.

We can continue this iterative process for every hidden unit,  strictly increasing the path kernel trace at each step, until $d_{in}=d_{out}=d$ for every hidden unit. 

After the completion of the previous process, every hidden unit $j$ will have the same number $d_j$ of input and output connections.
So, $2\times \sum_j d_j = m$. The path kernel trace for the entire network will then be:
\begin{equation}
    Tr(\bm{\Pi}_\theta) = \sum_{j=1}^n 2\, d_j \, K_{d_j}
\end{equation}
We can simplify this expression because $d_j=d$ for all $j$, based on Assumption-2.  
The resulting path kernel trace becomes:
\begin{equation}
    Tr(\bm{\Pi}_\theta) = 2n d \, K_{d} = m\, K_d
\end{equation}
The optimization problem can now be written as,
\begin{equation}
   d^* = \argmax_{d} \{m \times K_d\} \mbox{~such that~} d\leq D
\end{equation}
As $d$ increases so does the sum $K_d$ because $m$ is a constant. Therefore the  objective is maximized when $d$ takes the largest possible value $D$. That is, each of the $n$ hidden units will be connected to all of the input and output units, i.e., $d^* = D$.

So, the optimal width of the hidden layer is $n^* = m/(2D)$, which is the lowest possible width given the target number of edges $m$. In the next Lemma (A.2), we show that $P^* = n \, D^2$ is the maximum possible number of paths given $m$. 
 



\textbf{Open Question:} The previous result refers to a single hidden layer. We were not able to generalize it to deeper networks. If we simplify the problem by considering networks in which all edge-weights are equal to the same value, it is easy to show that the subnetwork that maximizes the number of input-output paths also maximizes the path kernel trace (see Appendix A.2).



\subsection{MLP architecture with maximum number of paths}\label{sec-numpaths}

Let us consider an ReLU MLP network with $L>1$ hidden layers, and with $N_l$ hidden layer units at layer $l=1,...,L$. Without loss of generality, we can consider the case that the number of both inputs and outputs is equal to $D$.

Let the number of connections in the un-pruned network be $M$, 
the target number of  remaining connections after pruning be $m$, and the target network density be $\rho=m/M$. 

For simplicity, assume that $m$ is such that we can form a fully-connected network with the same number $k$ of hidden units at each layer. In other words, we assume that there is an integer $k$ that satisfies the following equation:
$m = k \, (2D+(L-1)k)$.

The next Lemma identifies the pruned architecture with the maximum number of paths, given $m$.

\textbf{Lemma 2:} \textit{In an MLP network with a target number of parameters $m$, the maximum number of input-output paths  results when each hidden layer has the same number of units after pruning, and those units are fully connected with the units of the previous layer.}

\textbf{Proof:} 


\textbf{Base case:}   Consider a network with two hidden layers. Let the number of units in the two hidden layers of the unpruned network be $N_1$ and $N_2$. 

We first show that, for a hidden layer $l$, the number of paths through that layer is maximized by selecting $n_l\leq N_l$ fully-connected units -- and pruning all other units. Then we show that the number of paths is maximized when the two hidden layers have the same number of  (fully-connected) units after pruning, i.e., $n_1=n_2$.

Let the number of incoming and outgoing connections for unit $i$ in a hidden layer be $\{d_i,k_i\}$. Then the number of paths from the previous layer to the next layer is given by $P = \sum_{i=1}^{N_l} d_i\times k_i$.

Suppose we can identify two hidden units $X$ and $Y$ that are not fully-connected. The number of paths through $X$ from the previous layer to the next layer is $P(X) = d_x \times k_x$ -- and similarly for $Y$, $P(Y) = d_y \times k_y$. If $P(Y)>P(X)$ we move an edge from $X$ to $Y$ such that the edge is selected from either $d_x$ or $k_x$, choosing the larger among the two. Similarly, the edge is added to either $d_y$ or $k_y$, choosing the smaller among the two. This process causes the number of paths to strictly increase, that is, $\Delta P = max\{d_y,k_y\} - min\{d_x,k_x\} > 0$. In the case of a tie, if $P(X)=P(Y)$, we select the unit with the higher number of incoming/outgoing edges so that $max\{d_y,k_y\} \geq min\{d_x,k_x\}$. If there are still ties, we break them moving edges from lower-index units to higher-index units.

We repeat this process, increasing the number of paths in each iteration, until we cannot find any units that are not fully-connected. Then, the layer $l$ will consist of $n_l \leq N_l$ fully-connected units. 

We want to choose $n_1$ and $n_2$ to maximize the total number of paths:
\begin{equation}
     D^2\times n_1 \times n_2, \mbox{~such that~} m = D( n_1 +  n_2 ) + n_1 n_2 
\end{equation}
Substituting $n_2$, we want to maximize the function of $n_1$: 
\begin{equation}
   D^2 \times n_1 \times \dfrac{m-Dn_1}{D+n_1}
\end{equation}
The maximum results when $n_1 = \sqrt{D^2+m}-D$. Solving for $n_2$, we get that the corresponding with of layer-2 is $n_2 = \sqrt{D^2+m}- D = n_1$. Therefore, the number of paths is maximized by having the same number of units $n = n_1 = n_2$ in the two hidden layers.

\textbf{Inductive step:} The induction hypothesis on a network with $L-1$ hidden layers is: given the target number of connections $m$, the number of paths for a network with $L-1$ hidden layers is maximized by considering a fully-connected network with the same number of units in each hidden layer, $n\leq N_i, i=1,...,L-1$, and $m = n(2D+(L-2)n)$.

Consider now a network with $L$ hidden layers. 
From the induction hypothesis we know that the number of paths $P_{L-1}$ till the last hidden layer is maximized when each hidden layer contains the same number of units $n$ and is fully-connected. 
All the units in layer $L-1$ have the same number of incoming paths $P_{L-1}/n$.

Suppose that the number of selected units in the last hidden layer is $n_{L}\leq N_{L}$.
The number of paths $P = \sum_{i=1}^{n} P_{i}\times P_{L-1}/n = P_{L-1}/n\times \sum_{i=1}^{n} P_{i}$, where $P_i$ is the number of paths from the $i^{th}$ unit in hidden layer $L-1$ to the output units. 
Using the same approach with the base case, we see that $P$ is maximized when $n_{L} = n$.

\begin{figure*}
\centering
        \centering
         \begin{subfigure}[b]{0.245\textwidth}
             \centering
             \includegraphics[width=\textwidth]{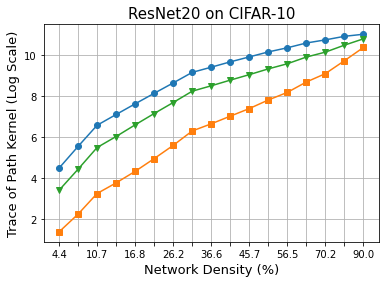}
         \end{subfigure}
           \centering
         \begin{subfigure}[b]{0.245\textwidth}
             \centering
             \includegraphics[width=\textwidth]{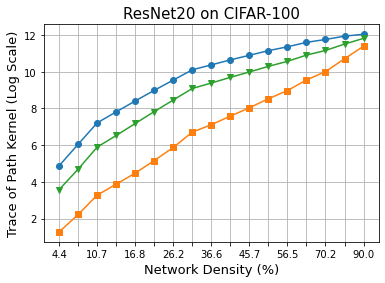}
         \end{subfigure}
         \centering
         \begin{subfigure}[b]{0.245\textwidth}
             \centering
             \includegraphics[width=\textwidth]{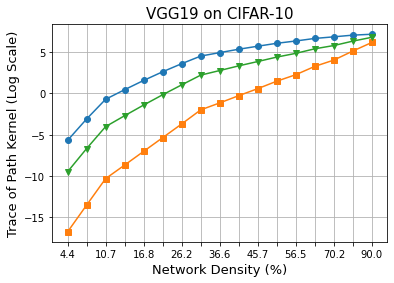}
         \end{subfigure}
           \centering
         \begin{subfigure}[b]{0.245\textwidth}
             \centering
             \includegraphics[width=\textwidth]{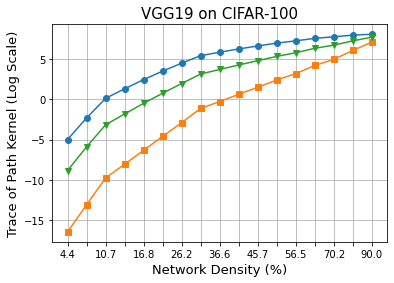}
         \end{subfigure}
           \centering
         \begin{subfigure}[b]{0.5\textwidth}
             \centering
             \includegraphics[width=\textwidth]{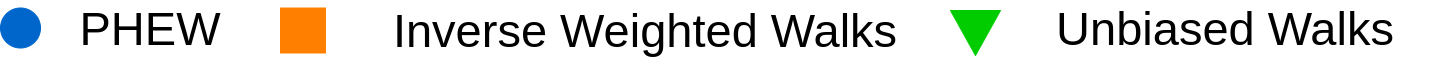}
         \end{subfigure}
    \caption{Comparison of the path kernel trace of sparse networks obtained using PHEW, unbiased random walks, and inverse-weighted random walks.}
    \label{pktracewalks}
\end{figure*}

\subsection{PHEW and per-layer width}\label{app:width}

Consider a fully-connected MLP network, $f:\R^D\rightarrow \R^K$, with $L$ hidden layers and $N_l$ units per layer
($l=1\dots L$). The weights are initialized with the Kaiming method, i.e., the initial weight from unit $j$ at layer $l-1$ to unit $i$ at layer $l$ is $\theta^{[l]}(i,j) \thicksim \mathcal{N}(0,\sigma_l^2)$, where $\sigma_l^2 = 2/N_l$.

\textbf{Lemma 3:}  \textit{The expected number of PHEW random walks through each unit of layer $l$ is $W/N_l$, where $W$ is the required number of walks to achieve the target network density $\rho$.}
In other words, at least in expectation, PHEW utilizes every hidden unit of a layer, resulting in the maximum per-layer width.

\textbf{Proof:} Let us first consider a network with a single hidden layer that has $N$ hidden units. 
In PHEW, we start the same number of walks  from each input unit.
So, given the number of walks required to achieve the target density $W$, the number of walks starting from any input unit is
$W/D$.

Let $W_n$ be the number of walks passing through hidden unit $n$. Then,
\begin{equation}
    \mathbb{E}(W_n) = \sum_{i=1}^D \dfrac{W}{D} \times Q^{[1]}(n,i)
\end{equation}
where $Q^{[1]}(n,i)$ is the probability that the walk will move from input unit $i$ to a unit $n$ at the first hidden layer. For PHEW random walks this expectation is given by,
\begin{equation}
    \mathbb{E}(W_n) = \dfrac{W}{D} \sum_{i=1}^D  \dfrac{|\theta^{[1]}(n,i)|}{\sum_{j=1}^N |\theta^{[1]}(j,i)|}
\end{equation}
The denominator of the previous equation is:
\begin{equation}
    \sum_{j=1}^N |\theta^{[1]}(j,i)| = N \times \left[ \dfrac{1}{N} \sum_{j=1}^N |\theta^{[1]}(j,i)|\right] = N\times \bar{\theta}_N(i)
\end{equation}
where
 $\bar{\theta}_N(i)$ is the sample mean of the folded normal distribution. 
For a sufficiently large sample size $N$ (in practice $>40$), the sample mean is approximately equal to the population mean $\mu$. So we replace $\bar{\theta}_N(i)$ with $\mu$ for all $i=1,...,D$:
\begin{equation}
    \mathbb{E}(W_n) \approx \dfrac{W}{N \mu} \, \dfrac{1}{D} \sum_{i=1}^D  |\theta^{[1]}(n,i)|
\end{equation}
Similarly we approximate the sample average of the previous equation with $\mu$:
\begin{equation}
    \mathbb{E}(W_n) \approx \dfrac{W}{N \mu} \mu = \dfrac{W}{N}
\end{equation}

Hence, the expected number of PHEW walks through any hidden unit is the same. 

Similarly, the number of walks through each of the $K$ output units is $W/K$, where $K$ is the number of output units.



For networks with more than one hidden layer, it is simple to use induction and show  that the expected number of random walks using PHEW through any unit at layer $l$ is $W/N_l$.

\subsection{PHEW and path kernel trace}\label{phewpktrace}

Consider a fully-connected MLP network, $f:\R^D\rightarrow \R^K$ with $L$ hidden layers and $N_l$ units per layer
($l=1\dots L$), and suppose that the weights are initialized using the Kaiming method.

\textbf{Lemma 4:} \textit{Consider two input-output paths $u$ and $b$ in the previous MLP network: $u$ has been selected with a uniform random walk, while $b$ has been selected with the PHEW random walk process. Then,}
\begin{equation}
    \mathbb{E}[\bm{\Pi_\theta}(b,b)] = 2^{L} \times \mathbb{E}[\bm{\Pi_\theta}(u,u)]
\end{equation}

\textbf{Proof:} The path kernel matrix is given as :
\begin{equation}
\bm{\Pi_\theta}(p, p') = \sum_{i=1}^m \left(\dfrac{\pi_p(\bm{\theta})}{\theta_{i}}\right)\left(\dfrac{\pi_{p'}(\bm{\theta})}{\theta_{i}}\right) p_i p'_i
\end{equation}

Let path $p$ be a path selected through a random walk process from an input unit to an output unit.
The diagonal element that represents the contribution of $p$ to the path-kernel trace is given by
\begin{equation}
    \bm{\Pi_\theta}(p, p) = \sum_{i=1}^m \left(\dfrac{\pi_p(\bm{\theta})}{\theta_{i}}\right)^2 p_i 
\end{equation}

Suppose that path $p$ is formed by the edge weights $\{\theta_p^{[l]}\}_{l=1}^{L+1}$. Then, 
\begin{equation}
    \bm{\Pi_\theta}(p, p) = \sum_{l=1}^{L+1} \left(\dfrac{\pi_p(\bm{\theta})}{\theta_p^{[l]}}\right)^2
\end{equation}
We can observe that $\bm{\Pi_\theta}(p, p)$ is a linear function with respect to  $\{(\theta_p^{[l]})^2\}_{l=1}^{L+1}$. Therefore,
\begin{equation}
    \mathbb{E}[\bm{\Pi_\theta}(p,p)] = \sum_{l=1}^{L+1} \left[\prod_{i=1,i\neq l}^{L+1} \mathbb{E}[\theta_p^{[i]}]^2\right]
\end{equation}

 The probability of selecting unit $j$ at layer $l$ given that the walk is on a unit $k$ at layer $l-1$ is $Q^{[l]}(j,k)$.

We compute the expected value of each squared-weight in a sequential manner. That is, 
\begin{equation}
    \mathbb{E}[(\theta_p^{[l]})^2] = \sum_{j=1}^{N_l} (\theta^{[l]}(j,k))^2 Q^{[l]}(j,k)
\end{equation}

\textbf{Uniform random walks:} First, suppose that the random walk is not biased. Denote an unbiased path by $u$. Then, $Q^{[l]}(j,k) = 1/N_l$.
Let us now start the walk with a randomly selected input unit $k$, 
\begin{equation}
    \mathbb{E}[(\theta_u^{[1]})^2] = \sum_{j=1}^{N_1} (\theta^{[1]}(j,k))^2 \dfrac{1}{N_1} \approx \sigma_1^2
\end{equation}
This is the sample mean of  $N_l$ squares from a normal distribution. If $N_l$ is sufficiently large, we can approximate the sample mean with the expected value $\mathbb{E}[(\theta^{[1]}(j,k))^2]$. 
Note that we are only interested in the expectation of the weight value and not the position of the corresponding connection. 

Moving to the next hop, we observe that the initialized weight distribution for all edges is the same regardless of the previous unit. Hence, 
\begin{equation}
    \mathbb{E}[(\theta_p^{[l]})^2] = \sigma_l^2
\end{equation}
Plugging the values back in the expectation of the path kernel trace,
\begin{equation}
    \mathbb{E}[\bm{\Pi_\theta}(u,u)] = \sum_{l=1}^{L+1} \left[\prod_{i=1,i\neq l}^{L+1} \sigma_l^2\right] 
\end{equation}

\textbf{PHEW random walks:} Now consider a path $b$ that is sampled using the PHEW biased random walk process. 
Here,
\begin{equation}
    Q^{[l]}(j,k) = \dfrac{|\theta^{[l]}(j,k)|}{\sum_{i=1}^{N_l}|\theta^{[l]}(i,k)|}
\end{equation}
Let us again start the random walk process with some input unit $k$,
\begin{equation}
\begin{aligned}
    \mathbb{E}[(\theta_b^{[1]})^2]& = \sum_{j=1}^{N_1} (\theta^{[1]}(j,k))^2 \dfrac{|\theta^{[l]}(j,k)|}{\sum_{i=1}^{N_1}|\theta^{[l]}(i,k)|}\\
    & = \dfrac{\sum_{j=1}^{N_1} |\theta^{[l]}(j,k)|^3}{\sum_{i=1}^{N_1}|\theta^{[l]}(i,k)|}\\
    & = \dfrac{\dfrac{1}{N_1} \sum_{j=1}^{N_1} |\theta^{[l]}(j,k)|^3}{\dfrac{1}{N_1} \sum_{i=1}^{N_1}|\theta^{[l]}(i,k)|}
\end{aligned}
\end{equation}

We approximate this ratio of sample means by the ratio of the corresponding two population means,  
\begin{equation}
    \mathbb{E}[(\theta_b^{[1]})^2] \approx \dfrac{2\sqrt{\dfrac{2}{\pi}}\sigma_1^3N_1}{\sqrt{\dfrac{2}{\pi}}\sigma_1N_1} = 2 \sigma_1^2 
\end{equation}
where the numerator is the expected value of $|X|^3, X\thicksim \mathcal{N}(0,\sigma_1^2)$, and the denominator is the expected value of $|X|, X\thicksim \mathcal{N}(0,\sigma_1^2)$.

\begin{figure*}
\centering
        \begin{subfigure}[b]{1\textwidth}
             \centering
             \includegraphics[width=\textwidth]{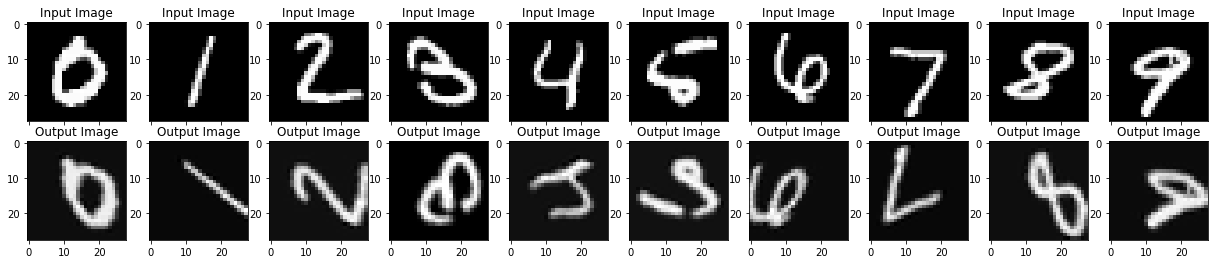}
         \end{subfigure}
\caption{Examples of inputs and outputs for the MNIST image transformation task}
    \label{fig:10}
\end{figure*}

Moving on to the next hop, we perform the same process as in the case of unbiased random walks, where the expected weight value is only dependent on the initialization distribution of the edges. We know that the initialization distribution is the same for all edges regardless of the unit they are connected to in the previous layer. Hence, 
\begin{equation}
    \mathbb{E}[(\theta_b^{[l]})^2] \approx 2 \sigma_l^2 
\end{equation}
Plugging this value back into the expected path kernel trace,
\begin{equation}
    \mathbb{E}[\bm{\Pi_\theta}(b,b)] = \sum_{l=1}^{L+1} \left[\prod_{i=1,i\neq l}^{L+1} 2 \sigma_l^2 \right] = 2^L \sum_{l=1}^{L+1} \left[\prod_{i=1,i\neq l}^{L+1} \sigma_l^2\right] 
\end{equation}

 We conclude that a path selected using PHEW will contribute to the path kernel trace much more than a path selected through an unbiased random walk, and this gap increases exponentially with the depth $L$ of the network:
\begin{equation}
    \mathbb{E}[\bm{\Pi_\theta}(b,b)] = 2^L \times \mathbb{E}[\bm{\Pi_\theta}(u,u)]
\end{equation}

\section{Image transformation task}\label{B}
In this section we present results for a regression task in which the number of inputs is the same with the number of outputs.
The task is to perform MNIST image transformations (rotation and shearing). The angle of rotation and the shearing coefficient are class-specific. We crop and pad the rotated image to fit the original dimensions. The classes are rotated in multiples of 30$^\circ$ and sheared with coefficients uniformly sampled between -0.6 and 0.6. Figure \ref{fig:10} shows various input-output examples.

In this task, we use MLPs with width equal to 100, 200, 300 and 400.  We use 1000 examples per class for training and 20 epochs.
The learning rate is 0.001 with an exponential decay factor of 0.95 after every epoch, Adam optimizer, batch size of 32. The loss function is the mean-squared error between the actual output and correct output. Test performance is evaluated using the MSE metric on the entire test set. ReLU in combination with batch-normalization are implemented at each hidden layer. 

\begin{figure*}
\centering
        \centering
         \begin{subfigure}[b]{0.245\textwidth}
             \centering
             \includegraphics[width=\textwidth]{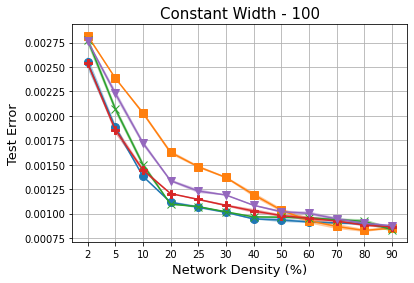}
         \end{subfigure}
           \centering
         \begin{subfigure}[b]{0.245\textwidth}
             \centering
             \includegraphics[width=\textwidth]{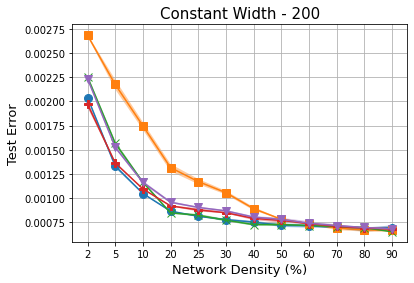}
         \end{subfigure}
         \centering
         \begin{subfigure}[b]{0.245\textwidth}
             \centering
             \includegraphics[width=\textwidth]{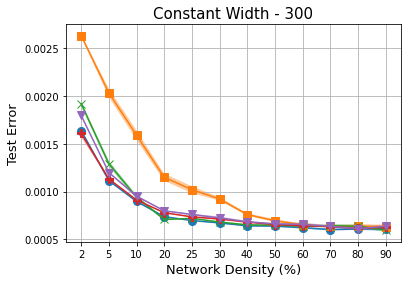}
         \end{subfigure}
           \centering
         \begin{subfigure}[b]{0.245\textwidth}
             \centering
             \includegraphics[width=\textwidth]{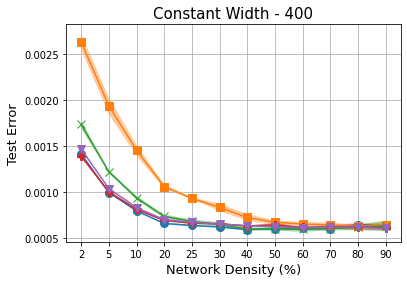}
         \end{subfigure}
           \centering
         \begin{subfigure}[b]{0.5\textwidth}
             \centering
             \includegraphics[width=\textwidth]{Figures/Main/Figure1/label.png}
         \end{subfigure}
    \caption{Comparison of PHEW against other pruning-before-training methods for the MNIST image transformation task.}
    \label{ComparisonTransformation}
\end{figure*}

\subsection{Comparison between PHEW and other pruning methods on transformation task}

Here, we present results for the image transformation task. Specifically, we compare the performance of PHEW with the two data-dependent methods SNIP and GraSP as well as with the data-agnostic methods SynFlow and SynFlow-L2.

Figure \ref{ComparisonTransformation} 
shows that PHEW performs better than the other algorithms in a wide range of network density values. Of course, as the density increases the differences between these methods diminish. 

An interesting observation here is that the data-dependent methods SNIP and GraSP perform worse in this task than the data-agnostic methods. 
This may be due to the change in the loss function from cross entropy to mean squared error for regression.
We plan to investigate this phenomenon further by introducing different tasks/loss functions and architectures.



\begin{figure*}
\centering
        \centering
         \begin{subfigure}[b]{0.245\textwidth}
             \centering
             \includegraphics[width=\textwidth]{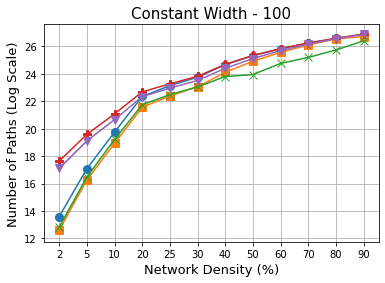}
         \end{subfigure}
           \centering
         \begin{subfigure}[b]{0.245\textwidth}
             \centering
             \includegraphics[width=\textwidth]{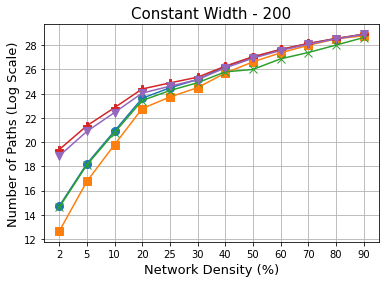}
         \end{subfigure}
         \centering
         \begin{subfigure}[b]{0.245\textwidth}
             \centering
             \includegraphics[width=\textwidth]{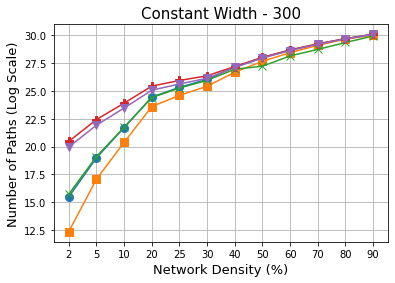}
         \end{subfigure}
           \centering
         \begin{subfigure}[b]{0.245\textwidth}
             \centering
             \includegraphics[width=\textwidth]{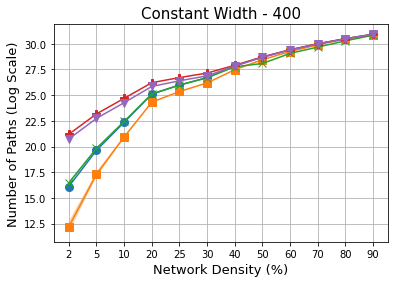}
         \end{subfigure}
           \centering
         \begin{subfigure}[b]{0.5\textwidth}
             \centering
             \includegraphics[width=\textwidth]{Figures/Main/Figure1/label.png}
         \end{subfigure}
    \caption{Comparison of the number of paths obtained through PHEW and other pruning methods. }
    \label{numpaths}
\end{figure*}
\subsection{Number of input-output paths}


Here, we present results for the number of input-output paths resulting from different pruning methods in the image transformation task -- see Figure \ref{numpaths}.

We  observe that the sub-networks resulting from SynFlow and SynFlow-L2 indeed have the highest number of input-output paths. 
This empirical observation provides further evidence for the connection between the magnitude of the path kernel trace and the number of input-output paths (see Appendix \ref{appmaxpkt}).

\begin{figure}
\centering
  \centering
         \begin{subfigure}[b]{0.2375\textwidth}
             \centering
             \includegraphics[width=\textwidth]{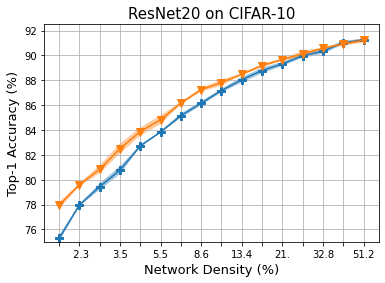}
         \end{subfigure}
        \centering
         \begin{subfigure}[b]{0.2375\textwidth}
             \centering
             \includegraphics[width=\textwidth]{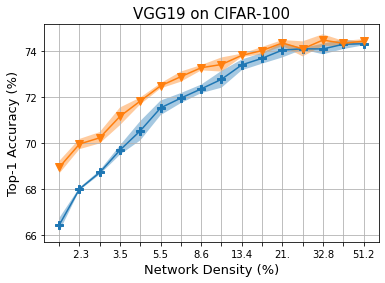}
         \end{subfigure}
           \centering
         \begin{subfigure}[b]{0.4\textwidth}
             \centering
             \includegraphics[width=\textwidth]{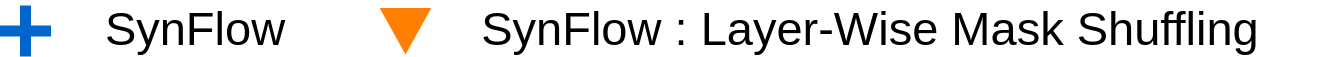}
         \end{subfigure}
    \caption{\textbf{SynFlow ablation:} Performance comparison of subnetworks obtained using SynFlow and subnetworks obtained by layer-wise randomized SynFlow subnetworks.}
    \label{sfablation}
\end{figure}

\section{Ablations of SynFlow and SynFlow-L2}\label{SF2ablations}

The formation of narrow subnetworks through SynFlow was first discovered in the ablation studies conducted in \cite{frankle2020pruning}. 
In that work, the edges of a layer are shuffled, creating a uniform distribution of connections across all units in the layer. 

\begin{figure}
\centering
  \centering
       \begin{subfigure}[b]{0.2375\textwidth}
             \centering
             \includegraphics[width=\textwidth]{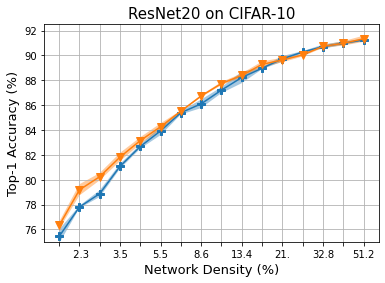}
         \end{subfigure}
        \centering
         \begin{subfigure}[b]{0.2375\textwidth}
             \centering
             \includegraphics[width=\textwidth]{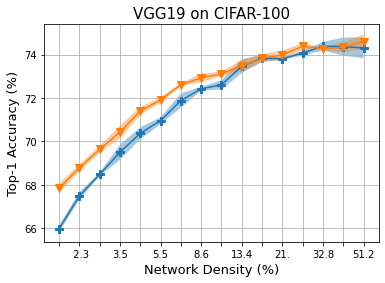}
         \end{subfigure}
           \centering
         \begin{subfigure}[b]{0.4\textwidth}
             \centering
             \includegraphics[width=\textwidth]{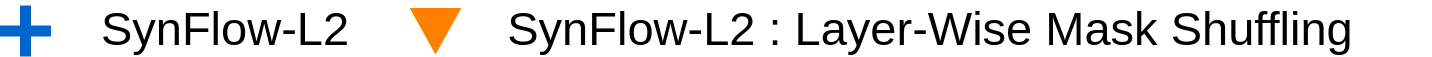}
         \end{subfigure}
    \caption{\textbf{SynFlow-L2 Ablation:} Performance comparison of subnetworks obtained using SynFlow-L2 and subnetworks obtained by random layer-wise weight shuffling of SynFlow-L2 subnetworks.}
    \label{sf2ablation}
\end{figure}

In Figures~\ref{sfablation} and
 \ref{sf2ablation}, we  observe that for both SynFlow and SynFlow-L2, random shuffling of weights  increases  performance. The performance increase becomes larger in datasets with more classes. We believe that this is due to the increased complexity of those datasets and the need for a larger layer width to learn disconnected decision regions \cite{nguyen2018neural}.  

The authors of \cite{frankle2020pruning} hypothesised that the cause of this effect is the iterative nature of the SynFlow algorithm along with the number of paths through specific units.
The scoring function for a specific edge-weight $\theta_i$ in SynFlow is $\sum_{p=1}^P |\pi_p(\bm{\theta})|p_i$. This represents the sum of edge-weight-products of all paths through $\theta_i$.
Similarly, for SynFlow-L2 the scoring function is $\sum_{p=1}^P \pi_p(\bm{\theta})^2p_i/|\theta_i|$.

In both of these iterative pruning methods, 
if a unit has some pruned edges in the first few iterations it becomes more likely to be pruned in subsequent iterations (due to the reduction in the number of paths that traverse that unit).
Hence, these iterative methods are more likely to  completely eliminate some hidden units. 
This conjecture is aligned with our observation that by maximizing the number of input-output paths, the SynFlow and SynFlow-L2 methods eliminate many hidden units and result in very narrow layers.

\section{Implementation details}\label{id}

In this section we report implementation details and hyperparameter values for our experiments.

\subsection{Networks, datasets and hyperparameters}
\begin{table*}[!t]
\scriptsize
\centering
\begin{center}
\begin{tabular}{c c c c c c c c c}

\hline

 \textbf{Network} & \textbf{Dataset} & \textbf{Epochs} & \textbf{Batch-size} & \textbf{Optimizer} & \textbf{Momentum} & \textbf{Learning Rate} & \textbf{Lr Drop} & \textbf{Weight Decay} \\ 
 \hline

ResNet20 & CIFAR-10 & 160 & 128 & SGD & 0.9 & 0.1 & 10x at 1/2 and 3/4 epochs & 1e-4\\
ResNet20 & CIFAR-100 & 160 & 128 & SGD & 0.9 & 0.1 & 10x at 1/2 and 3/4 epochs & 1e-4\\
VGG19 & CIFAR-10 & 160 & 128 & SGD & -- & 0.1 & 10x at 1/2 and 3/4 epochs & 5e-3\\
VGG19 & CIFAR-100 & 160 & 128 & SGD & -- & 0.1 & 10x at 1/2 and 3/4 epochs & 5e-3\\
ResNet18 & Tiny-ImageNet & 200 & 256 & SGD & 0.9 & 0.2 & 10x at 1/2 and 3/4 epochs & 1e-4\\
VGG19 & Tiny-ImageNet & 200 & 256 & SGD & 0.9 & 0.2 & 10x at 1/2 and 3/4 epochs & 1e-4\\
\hline

\end{tabular}
\end{center}
\caption{Hyperparameters used for various combinations of datasets and networks in the experiments.}
\label{tab:1}
\end{table*}
We present results on three networks and three datasets for classification. The combination of networks and datasets used are as follows:
\begin{enumerate}
    \item ResNet20 and CIFAR-10
    \item ResNet20 and CIFAR-100
    \item VGG19 and CIFAR-10
    \item VGG19 and CIFAR-100
    \item ResNet18 (Modified) and Tiny-ImageNet
    \item VGG19 and Tiny-ImageNet
\end{enumerate}

\textbf{ResNet20}: The ResNet20 architecture was designed for CIFAR-10 \cite{he2016deep}. The same network is also used for CIFAR-100. The network consists of 20 layers and batch-normalization is performed before activation.

\textbf{VGG19} : The VGG19 architecture was taken from the original paper \cite{simonyan2014very}. The network consists of five convolutional blocks followed by max-pooling. The first block consist of two layers with width 64 -- that is 64 filters. The second block contained two layers with width of 128; the third, fourth and fifth blocks consist of three layers each with the same width: 256, 512 and 512 respectively. All the filters have dimension $3\times 3$, with the number of channels varying. For CIFAR-10/100 we replace the three linear layers in the original network by a single layer with width equal to the number of classes. 

\textbf{ResNet18} (Modified): ResNet18 was originally proposed for classification on ImageNet \cite{he2016deep}. A modification of the original architecture was provided by the authors of SynFlow \cite{tanaka2020pruning} for Tiny-ImageNet. For a fair comparison we use the same architecture with the SynFlow paper. Specifically, the first convolutional layer uses a filter of dimensions $3\times 3$ with a stride of 1, without the use of max-pooling. 

All networks are initialized with Kaiming Normal initialization (\cite{he2015delving}). 

\textbf{Datasets:} We use three well known datasets for classification: CIFAR10, CIFAR-100 and Tiny-ImageNet, (the smaller version of ImageNet dataset)\cite{russakovsky2015imagenet}. Due to resource limitations we are unable to present results for the full ImageNet dataset. For CIFAR-10/100 we perform channel-wise normalization using mean and standard deviation and use random horizontal flipping. For Tiny-ImageNet we use channel-wise normalization, random cropping of size $64\times 64$ and random horizontal flipping.

\textbf{Hyperparameters:} We provide the hyperparameters  for each of the networks and datasets in Table \ref{tab:1}. Note that these hyperparameters were tuned for the unpruned network and not for sparse networks.

\subsection{Pruning mechanisms}

The pruning algorithms that we use as baselines follow a two-step process: first a score is computed for each edge, and then some edges are eliminated  based on this score. 
Below we report various implementation details for each baseline. 
We use the same training hyperparameters for every baseline. 


\textbf{Random pruning:} Upon initialization, all edges are assigned a score. The score follows a uniform distribution with 0-1 range. 

\textbf{Magnitude pruning after training:} First, the given dense network is trained.
Each edge is assigned a score that is equal to the absolute value of its weight. Then, the network is pruned iteratively, with one epoch of training (fine-tuning) after each pruning iteration.  Once the target network density is obtained, the sparse network is fully trained once again. Note that the weights are not reverted back to the initial weights for that second training cycle.   

We use the pruning schedule of \cite{zhu2017prune} for iterative pruning. The number of pruning iterations is five for ResNet20 on CIFAR-10, ten for VGG19 on CIFAR-100 and Tiny-ImageNet, and twentyfive for ResNet18 on   Tiny-ImageNet.

\textbf{SNIP:} Upon initialization, we compute the SNIP score $\left|\dfrac{\partial \bm{L}}{\partial \theta} \odot \theta \right|$ using a random subset of training data. That data is chosen such that they consist of 10 images per class \cite{lee2018snip}. The scores are computed for batches of size 256 for CIFAR-10/100 and 64 for Tiny-ImageNet and summed across  batches.

\textbf{GraSP:} Similar to SNIP but the score is given by  $\left(H \dfrac{\partial \bm{L}}{\partial \theta}\right) \odot \theta$.

\textbf{SynFlow:} Given the desired final network density, we compute the target density at each iteration of the algorithm.
Then we compute the SynFlow score given by, $\left|\dfrac{\partial \bm{R_{SF}}}{\partial \theta} \odot \theta \right|$ and prune the lowest scoring edges. 
We repeat this process over 100 iterations, with an exponential decay of the network density, to achieve the target density at the 100$^{th}$ iteration.

\textbf{SynFlow-L2:} Similar to SynFlow but the score is given by $\left|\dfrac{\partial \bm{R_{SF2}}}{\partial \theta^2} \odot \theta \right|$.

\textbf{PHEW:} We first compute the discrete probability distribution for all possible states in both forward and backward directions. 


The selection of the starting unit for each random walk in the forward (or backward) direction is performed in a round robin manner.

The random walk process is repeated until the target density is achieved. 



\end{document}